\documentclass{article}
\usepackage{spconf,amsmath,graphicx}
\usepackage{caption}
\usepackage{amssymb}
\usepackage{algorithm}
\usepackage{algpseudocode}
\usepackage{xcolor}
\usepackage{multirow}
\usepackage[bookmarks=false]{hyperref}
\usepackage{cite}
\usepackage{multicol}
\usepackage{xurl}

\usepackage{caption}
\makeatletter
\g@addto@macro\@maketitle{
  \begin{center}
  \includegraphics[width=.92\textwidth]{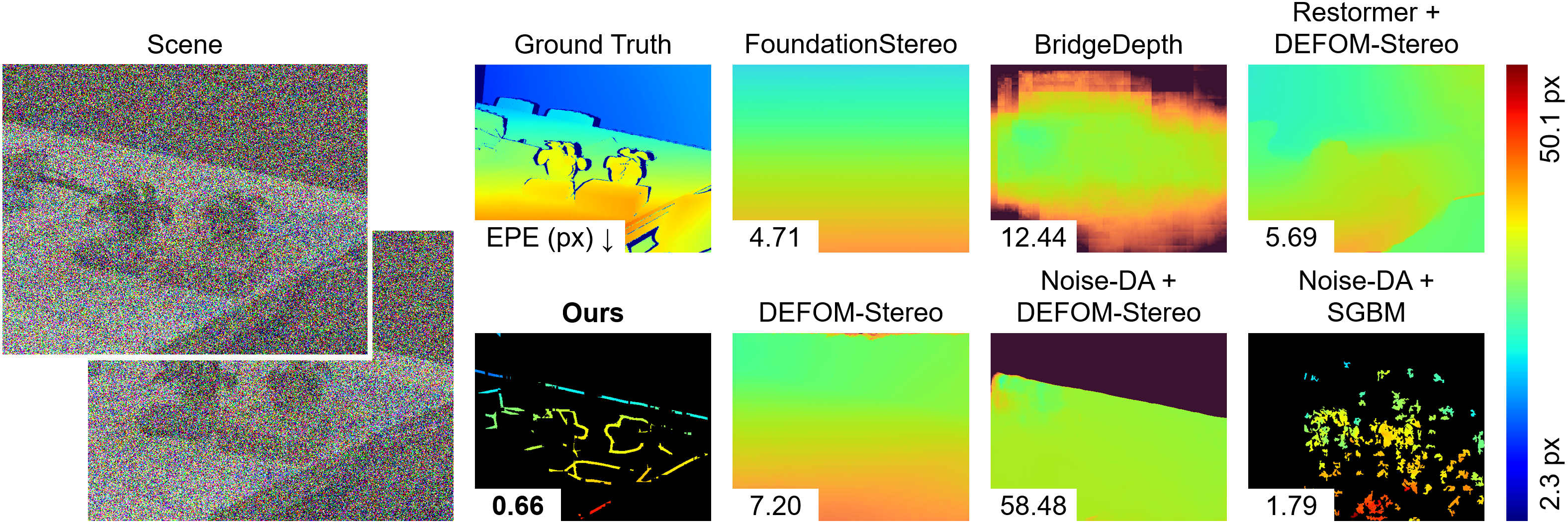}
  \captionof{figure}{Stereo matching under severe noise. Previous stereo algorithms~\cite{wen2025foundationstereo,jiang2025defom,guan2025bridgedepth,hirschmuller2007stereo} recover disparity maps by finding correspondences based mostly on fine image textures. However, under severe noise, these fine textures are badly degraded, causing existing methods to fail, even combined with denoising algorithms~\cite{liao2024denoising,Zamir2021Restormer}. Our approach instead extracts coarse visual features, which remain discernible under noise, and estimates disparity on the extracted image representations. This yields sparse disparity maps with substantially higher accuracy. Inset EPEs~\cite{scharstein2002taxonomy} (px) are calculated on unmasked disparity predictions.}
  \label{fig:teaser}
  \end{center}
  \vspace{0.15in}
}
\makeatother

\def\x{{\mathbf x}}

\title{Non-Learning Low-light Stereo Vision}

\name{Jason Wang$^{1*}$, Lucas Nguyen$^{2*}$, Hyunseung Eom$^{2*}$\thanks{$^{*}$Equal contribution.}, Wei Xu$^{2}$, and Qi Guo$^{2\dag}$\thanks{$^{\dag}$Corresponding author: \href{mailto:qiguo@purdue.edu}{qiguo@purdue.edu}.}}
\address{$^{1}$Department of Computer Sciences, Purdue University\\
$^{2}$Elmore Family School of Electrical and Computer Engineering, Purdue University}

\begin{document}

\maketitle

\begin{abstract}
We present a non-learning stereo framework for disparity estimation from severely noisy images. Using the Field of Junctions (FoJ)~\cite{verbin2021foj}, it retains coarse visual features stable under severe noise for cost volume construction while discarding fine textures inseparable from photon noise. The resulting structural information guides boundary-aware Semi-Global Matching (SGM)~\cite{hirschmuller2005accurate} that dynamically adapts smoothness penalties to preserve true disparity discontinuities. The output is a sparse disparity map more accurate than those of recent stereo algorithms over unmasked pixels on widely-used benchmark datasets. 
\end{abstract}

\section{Introduction}
\label{sec:intro}
Stereo vision is widely used in 3D reconstruction~\cite{geiger2011stereoscan}, autonomous navigation~\cite{geiger2013vision}, and robotic perception~\cite{shi2024asgrasp}. However, estimating disparities from photon-limited stereo images remains challenging because severe noise corrupts visual features, making reliable keypoint detection and correspondence difficult~\cite{huang2022low}. Existing pipelines that cascade a denoiser with a stereo matcher often introduce view-inconsistent details, which disrupt feature matching~\cite{10.1007/978-3-030-01228-1_7}.

Complementary to this line of work, we propose a non-learning framework that extracts only coarse visual features (e.g., long edges and large blobs) from the stereo image pair—features that remain discernible under noise, unlike fine details. These extracted representations are then fed into a stereo matcher. The framework produces sparse disparity estimates along boundaries of coarse structures that are more accurate than those from previous stereo or denoiser+stereo algorithms, as illustrated in Fig.~\ref{fig:teaser}. This demonstrates the effectiveness of our framework in photon-limited settings.

Specifically, we utilize the Field-of-Junctions (FoJ)~\cite{verbin2021foj}, which converts the input images into parametric representations of overlapping image patches. Prior work has shown that parametric local image representations, such as FoJ, are remarkably robust in extracting coarse visual features under severe noise~\cite{verbin2021foj, xu2024ctbound, miaboundary, xu2025blurry}.  Leveraging the FoJ structures together with the original noisy images, we construct a bi-directional cost volume and estimate disparities for each FoJ patch using Semi-Global Matching (SGM) with a modified boundary-aware penalty term~\cite{hirschmuller2005accurate}.  

The contributions of this work are summarized as follows:
\begin{enumerate}
    \item A novel non-learning framework for joint structure discovery and stereo matching that leverages the FoJ representation to construct robust cost volumes;
    \item An adapted SGM formulation with dynamically weighted smoothness penalties for reliable disparity estimation on sparse structures;
    \item A comprehensive experimental evaluation demonstrating superior accuracy over recent stereo and denoising-based stereo approaches under severe noise.
\end{enumerate}

All code and data of this work can be accessed at \textcolor{blue}{\url{https://github.com/guo-research-group/nonlearning-lowlight-stereo}}.

\section{Related Work}
\label{sec:related works}

\paragraph*{Analytical Stereo Depth Estimation.}
Classical stereo depth estimation relies on geometric constraints between two calibrated cameras to infer scene depth from pixel disparities. Early analytical methods formulated this as a correspondence problem, seeking to match image patches across the left and right views. A foundational approach is the sliding window method~\cite{kanade2002stereo}, which compares small local regions using metrics such as the sum of absolute differences (SAD), normalized cross-correlation (NCC), or census transform. Although efficient, window-based matching often fails near occlusions or textureless areas, where the assumption of locally constant disparity is violated.

To improve upon purely local methods, SGM~\cite{hirschmuller2005accurate} introduced an energy minimization framework that aggregates matching costs along multiple 1D paths while enforcing smoothness constraints across the image. SGM remains one of the most widely adopted analytical stereo algorithms due to its balance between accuracy and computational efficiency. Subsequent works have refined these formulations with adaptive support weights, sub-pixel interpolation, and cost regularization to better handle depth discontinuities and low-texture regions. However, these analytical approaches are inherently sensitive to image noise and depend heavily on reliable photometric similarity, limiting their performance under photon-limited or defocused imaging conditions.

\paragraph*{Deep Learning for Stereo Depth Estimation.}
The advent of deep learning has led to major advances in stereo depth estimation through end-to-end convolutional neural networks (CNNs) and cost volume aggregation. Methods such as DispNet~\cite{mayer2016large} first demonstrated the feasibility of learning disparity directly from synthetic stereo data, while GC-Net~\cite{kendall2017end} and PSMNet~\cite{chang2018pyramid} introduced 3D convolutional cost volumes to jointly reason about spatial and disparity context. Later architectures, including GANet~\cite{zhang2019ga}, GWCNet~\cite{guo2019group}, and LEAStereo~\cite{cheng2020hierarchical}, further improved precision near boundaries and fine structures through hierarchical aggregation, attention mechanisms, and adaptive correlation.

Despite their success, deep stereo networks often degrade when confronted with image noise, blur, or exposure variations, since they are trained on high-quality datasets such as Scene Flow~\cite{mayer2016large} or KITTI~\cite{geiger2013vision} that lack realistic noise distributions. To mitigate this, recent works have explored domain adaptation, uncertainty modeling, and self-supervised fine-tuning under real-world conditions. However, these networks still depend on strong texture cues, making them fragile under photon-limited or defocused settings where traditional matching signals collapse.

\paragraph*{Coarse Structure Based Stereo Methods.}
Recent methods that focus on high-level structural cues have shown impressive robustness under photon-limited conditions. Sharma and Cheong~\cite{10.1007/978-3-030-01228-1_7} presented an iterative framework that jointly optimizes structure and disparity by decomposing stereo matching into two subproblems: structure discovery and stereo matching. They demonstrate that their structure-based approach significantly outperforms baseline denoise-then-match frameworks. This approach, along with FoJ~\cite{verbin2021foj}, justifies the use of coarse structural priors to regularize noisy inputs. Our work builds upon these insights by using FoJ as a fundamental representation that provides both denoising and structure extraction necessary for stereo matching under photon-limited conditions.

\section{Methods}

\begin{figure}[t]
    \centering
    \includegraphics[width=.92\columnwidth]{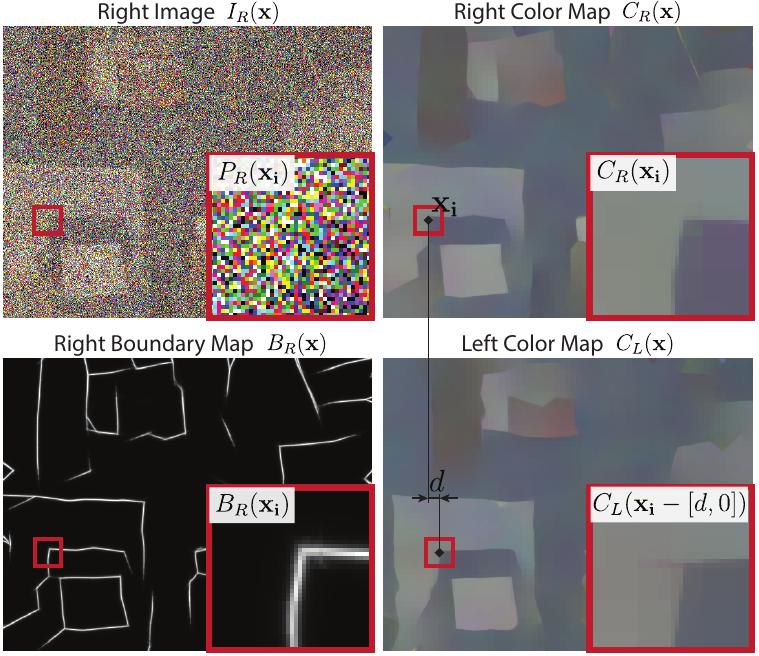}
    \caption{Color and boundary maps. Given a noisy image (top left), Field-of-Junctions (FoJ) extract the coarse visual features to generate the color map (top right) and the boundary map (bottom left). The proposed framework then constructs the cost volume for stereo matching using the color maps and the original stereo image pairs. }
    \label{fig:cost_volume}
\end{figure}

Given an image $I(\x)$, its Field-of-Junctions (FoJ) representation is a set of vectors, $\Pi =\{\boldsymbol{\theta}(\x_i)\}_{i=1}^N$. Each vector $\boldsymbol{\theta}(\x_i)$ represents the structure of the uniform-dimension image patch $P(\x_i)$, centered at $\x_i$, as a vertex and a fixed, pre-defined number of colored wedges. By averaging the FoJ representations of all patches, a global boundary map $B(\x)$ and a color map $C(\x)$ can be generated. The boundary map indicates the location of edges. The color map preserves the major image structures while attenuating noise in the original image~\cite{verbin2021foj}. Figure~\ref{fig:cost_volume} shows the boundary map and the color map extracted from a sample image.

\paragraph*{Initial Field-of-Junctions (FoJ) Representation Generation.} Given a rectified, noisy stereo image pair, $I_L(\x)$ and $I_R(\x)$, we first utilize the non-learning approach in Verbin et al.~\cite{verbin2021foj} to extract their FoJ representations of each image independently, denoted as $\Pi_L$ and $\Pi_R$. Sample color maps of a stereo image pair, $C_L(\x)$ and $C_R(\x)$, are shown in Fig.~\ref{fig:cost_volume}, which effectively preserves coarse image structures of the input images, such as long edges and large blobs, while attenuating inseparable fine visual structures and noise. 

\paragraph*{Cost Volume Reconstruction.} Our method predicts a disparity value for each image patch $P_{L/R}(\x_i)$. The left and right cost volumes are defined as:
\begin{equation}
    \begin{aligned}
    \mathbf{C}_{L/R}(\x_i, d) = &\lVert C_{L/R}(\x_i) - P_{R/L}(\x_i \pm [d,0]) \rVert^2 + \\
    &\lVert P_{L/R}(\x_i) - C_{R/L}(\x_i \pm [d,0]) \rVert^2 + \\
    \lambda_r &\lVert C_{L/R}(\x_i) - C_{R/L}(\x_i \pm [d,0]) \rVert^2,
\end{aligned}
\end{equation}
where $d$ denotes the candidate disparity, $C_{L/R}(\x_i)$ represents the patch centered at $\x_i$ of the color map corresponding to the left or right image, and the coefficient $\lambda_r$ balances the terms. This symmetric cost volume thoroughly considers matching across the color map and the original noisy image and between color maps of different views.

\paragraph*{Boundary Aware Semi-Global Matching (SGM).} Given the cost volumes $\mathbf{C}_{L/R}(\x_i,d)$, the left or right disparity map is generated using the standard SGM algorithm~\cite{hirschmuller2005accurate}:
\begin{equation}
    \begin{aligned}
    D_{L/R}(\x_i) = &
    \arg\min_{d}~ L_{\boldsymbol{r}}(\x_i, d), \text{~where} \\
    L_{\boldsymbol{r}}(\x_i, d) = ~&\mathbf{C}_{L/R}(\x_i, d) + \min \Bigg\{ L_{\boldsymbol{r}}(\x_i-\boldsymbol{r},d), \\
    &L_{\boldsymbol{r}}(\x_i-\boldsymbol{r},d-1) + P_1, \\
    &L_{\boldsymbol{r}}(\x_i-\boldsymbol{r},d+1) + P_1, \\
    &\min_\delta L_{\boldsymbol{r}}(\x_i-\boldsymbol{r},\delta) + P_2(\x_i -\boldsymbol{r}) \Bigg\} -  \\&\min_{k} L_r(\x_i-\boldsymbol{r},k).
\end{aligned}
\end{equation}
Here, the vector $\boldsymbol{r}$ denotes the eight aggregation directions. We propose to use a new $P_2$ term:
\begin{align}
    P_{2}(\x)=\max(P_{2, \text{base}} \exp(-\alpha B(\x)),P_{2, \text{min}}).
\end{align}
As $B(\x)\in[0,1]$ denotes the boundary strength at position $\x$, the term $P_2$ becomes small, i.e., $P_2\approx P_{2,\text{min}}$, when $\x$ is close to the boundary ($B(\x) \approx 1$), reducing the penalty on large disparity variations. 

\paragraph*{Post Processing.}
Following SGM, we upsample the per-patch disparity $D_{L/R}(\x_i)$ by assuming each patch shares constant disparity values and average the overlapping patches:
\begin{align}
    D_{L/R}(\x) = \mathbb{E}_{i, P_{L/R}(\x_i) \ni \x} \left[D_{L/R}(\x_i)\right].
\end{align}
We then filter the resulting disparity map $D_{L/R}(\x)$ according to the corresponding boundary map $B_{L/R}(\x)$ with a preset threshold $B_0$ and perform the standard Left-Right Consistency (LRC) check~\cite{hirschmuller2005accurate} to generate the final sparse disparity map:
\begin{equation}
    \begin{aligned}
    D^*(\x) = \text{LRC}\Big(&D_L(\x)[B_L(\x) > B_0], \\
    &D_R(\x)[B_R(\x) > B_0] \Big).
\end{aligned}
\end{equation}

\section{Results}

\paragraph*{Datasets and Preprocessing.}
For benchmarking against state-of-the-art methods, we evaluate our method on all 23 scenes of the  Middlebury 2014~\cite{scharstein2014high} dataset and on 89 scenes of the InStereo2K~\cite{bao2020instereo2k} dataset (comprising all 50 test scenes and the first 39 training scenes\footnote{None of the learning-based methods considered in the comparison is trained on this dataset.}). Prior to evaluation, we preprocess the stereo pairs in two distinct ways. First, images are uniformly downsampled to produce a lower-resolution test set compatible with our model’s input constraints. Second, to simulate photon-limited conditions, we apply the Poisson--Gaussian noise model via:
\begin{equation}
I(x) = \mathrm{Poisson}\!\left(\alpha I^{*}(x)\right) + \mathrm{Gaussian}\!\left(0, \sigma^{2}\right),
\label{eq:poisson_gaussian}
\end{equation}
where $I(x)$ and $I^{*}(x) \in [0,1]$ are the noisy and normalized clean images, respectively, $\alpha = 2$ is the photon level that controls the maximum photon capacity for each pixel, and $\sigma = 2$ is the standard deviation of the Gaussian read noise.

\paragraph*{Implementation Details.}
We use the same FoJ~\cite{verbin2021foj} optimization parameters as in the original FoJ unless otherwise specified. We set a patch size of $R=40$ pixels, with a stride of $8$ pixels. For our boundary-aware SGM implementation, we use $P_1=20$, $P_{2,\text{base}}=200$, $P_{2,\text{min}}=50$ and $\alpha=2$. For post-processing, we use a boundary threshold of $0.1$ with an LRC threshold of $1$ pixel. For Middlebury 2014, we use a block-based processing technique similar to Blurry-Edges~\cite{xu2025blurry}, as FoJ is memory-intensive for large images. We divide large images into overlapping blocks and estimate disparity for each block independently. Then, to reduce edge effects, we remove $N_{\text{margin}}$ patches from all sides of each block. We then aggregate all blocks to form the final full-size disparity estimation. For the evaluation on the Middlebury 2014 dataset, we use a block size of $(300,300)$ with $N_{\text{margin}}=4$.

\paragraph*{Evaluation.}

Since the predicted boundary location may differ from the ground truth (GT) boundary positions, we adopt a windowed evaluation method which tolerates boundary localization error. Each prediction is compared against the GT disparity values within a $W \times W$ neighborhood, and the lowest error is recorded. Throughout this paper, $W$ is set to $19$. We confirm in the supplement that our method's advantage is consistent across $W \in [1,19]$~px. All methods are evaluated under this protocol throughout the paper.

As our method produces a sparse disparity map retaining only high-confidence predictions, the metrics for our method are computed over predicted pixels only, representing approximately $2.8$\% coverage on Middlebury 2014 and $2.5$\% on InStereo2K. Competing methods that produce dense disparity maps are evaluated on all valid GT pixels, as they do not provide confidence masks to filter out unreliable predictions. 

\paragraph*{Quantitative Analysis.}
Table~\ref{tab:quantitative_comparison} reports quantitative stereo matching results using previous stereo or denoiser+stereo algorithms on noisy images generated from Middlebury 2014 and InStereo2K.  Our method achieves the lowest EPE and Bad-\textit{K} errors across both datasets. On Middlebury 2014, the second-best model (Noise-DA~\cite{liao2024denoising} + SGBM~\cite{hirschmuller2007stereo})  trails by approximately $0.7$~px in EPE, and $10$, $5$, and $3$ percentage points in Bad-1,3,5, respectively. On InStereo2K, our method improves upon the second-best method (Noise-DA~\cite{liao2024denoising} + SGBM~\cite{hirschmuller2007stereo}) by $0.6$ px in EPE, and approximately 18, 9, and 4 percentage points in Bad-1,3,5, respectively. The results clearly demonstrate the effectiveness of our method to automatically select discernible features to produce accurate disparity estimations from the severely contaminated images. 

\begin{table*}[tb!]
\centering
\caption{Quantitative comparison with SOTA methods on the Middlebury 2014 and InStereo2K datasets. Results show End Point Error (EPE)~\cite{scharstein2002taxonomy} and Bad-\textit{K} metrics~\cite{scharstein2002taxonomy} for $K={1,3,5}$ pixel thresholds evaluated with the window method. The proposed approach achieves the best performance across all metrics on both datasets.}
\label{tab:quantitative_comparison}
\begin{tabular}{@{}l@{\hskip 0.04in}|@{\hskip 0.06in}l@{\hskip 0.06in}|@{\hskip 0.06in}c@{\hskip 0.06in}|@{\hskip 0.1in}cccc@{}}
\hline
\multicolumn{2}{@{}l@{\hskip 0.06in}|@{\hskip 0.06in}}{\textbf{Method}} & \textbf{Venue} & \textbf{EPE (px)} $\downarrow$ & \textbf{Bad-1 (\%)} $\downarrow$ & \textbf{Bad-3 (\%)} $\downarrow$ & \textbf{Bad-5 (\%)} $\downarrow$ \\
\hline
\multirow{9}{*}{\rotatebox[origin=c]{90}{Middlebury 2014~\cite{scharstein2014high}}}

& Selective-IGEV~\cite{wang2024selective} & CVPR\textquotesingle24 & 27.53 & 94.60 & 89.96 & 85.38  \\
& BridgeDepth~\cite{guan2025bridgedepth} & ICCV\textquotesingle25 & 72.84 & 95.90 & 92.81  & 90.71  \\
& IGEV++~\cite{xu2025igev++} & TPAMI\textquotesingle25 & 34.02 & 91.33  & 83.29 & 76.98 \\
& DEFOM-Stereo~\cite{jiang2025defom} & CVPR\textquotesingle25 & 11.13 & 79.44 & 63.97 & 53.83 \\
& FoundationStereo~\cite{wen2025foundationstereo} & CVPR\textquotesingle25 & 12.80 & 81.43 & 67.78 & 58.12 \\
& Noise-DA~\cite{liao2024denoising} + SGBM~\cite{hirschmuller2007stereo} & ICLR\textquotesingle25\,+\,TPAMI\textquotesingle07 & 1.71 & 27.00  & 12.54  & 8.10  \\
& Restormer~\cite{Zamir2021Restormer} + DEFOM-Stereo~\cite{jiang2025defom} & CVPR\textquotesingle22\,+\,CVPR\textquotesingle25 & 5.47 & 64.88 & 44.55 & 32.34 \\
& Noise-DA~\cite{liao2024denoising} + DEFOM-Stereo~\cite{jiang2025defom} & ICLR\textquotesingle25\,+\,CVPR\textquotesingle25 & 6.72 & 71.29 & 52.05 & 38.72 \\

& \textbf{Ours} & -- & \textbf{0.98} & \textbf{17.30} & \textbf{7.88} & \textbf{5.03}\\
\hline
\multirow{11}{*}{\rotatebox[origin=c]{90}{InStereo2K~\cite{bao2020instereo2k}}}

& Selective-IGEV~\cite{wang2024selective} & CVPR\textquotesingle24 & 38.17 & 98.93  & 97.86  & 96.32  \\
& BridgeDepth~\cite{guan2025bridgedepth} & ICCV\textquotesingle25 & 88.12 & 98.24  & 96.69  & 95.58  \\
& IGEV++~\cite{xu2025igev++} & TPAMI\textquotesingle25 & 64.37 & 98.14  & 95.84  & 93.80  \\
& DEFOM-Stereo~\cite{jiang2025defom} & CVPR\textquotesingle25 & 5.06 & 77.41  & 52.28  & 33.75  \\
& FoundationStereo~\cite{wen2025foundationstereo} & CVPR\textquotesingle25 & 8.30 & 70.65  & 47.99  & 34.65  \\
& Noise-DA~\cite{liao2024denoising} + SGBM~\cite{hirschmuller2007stereo} & ICLR\textquotesingle25\,+\,TPAMI\textquotesingle07 & 2.10 & 50.95  & 20.59  & 10.23  \\
& Restormer~\cite{Zamir2021Restormer} + DEFOM-Stereo~\cite{jiang2025defom} & CVPR\textquotesingle22\,+\,CVPR\textquotesingle25 & 6.88 & 54.54  & 30.55  & 20.75  \\
& Noise-DA~\cite{liao2024denoising} + DEFOM-Stereo~\cite{jiang2025defom} & ICLR\textquotesingle25\,+\,CVPR\textquotesingle25 & 33.17 & 66.59  & 43.51  & 32.93  \\
& \textbf{Ours}  & -- & \textbf{1.50} & \textbf{33.21} & \textbf{11.10} & \textbf{6.68 } \\
\hline
\end{tabular}
\end{table*}

\begin{figure*}[h!]
    \centering
    \includegraphics[width=\textwidth]{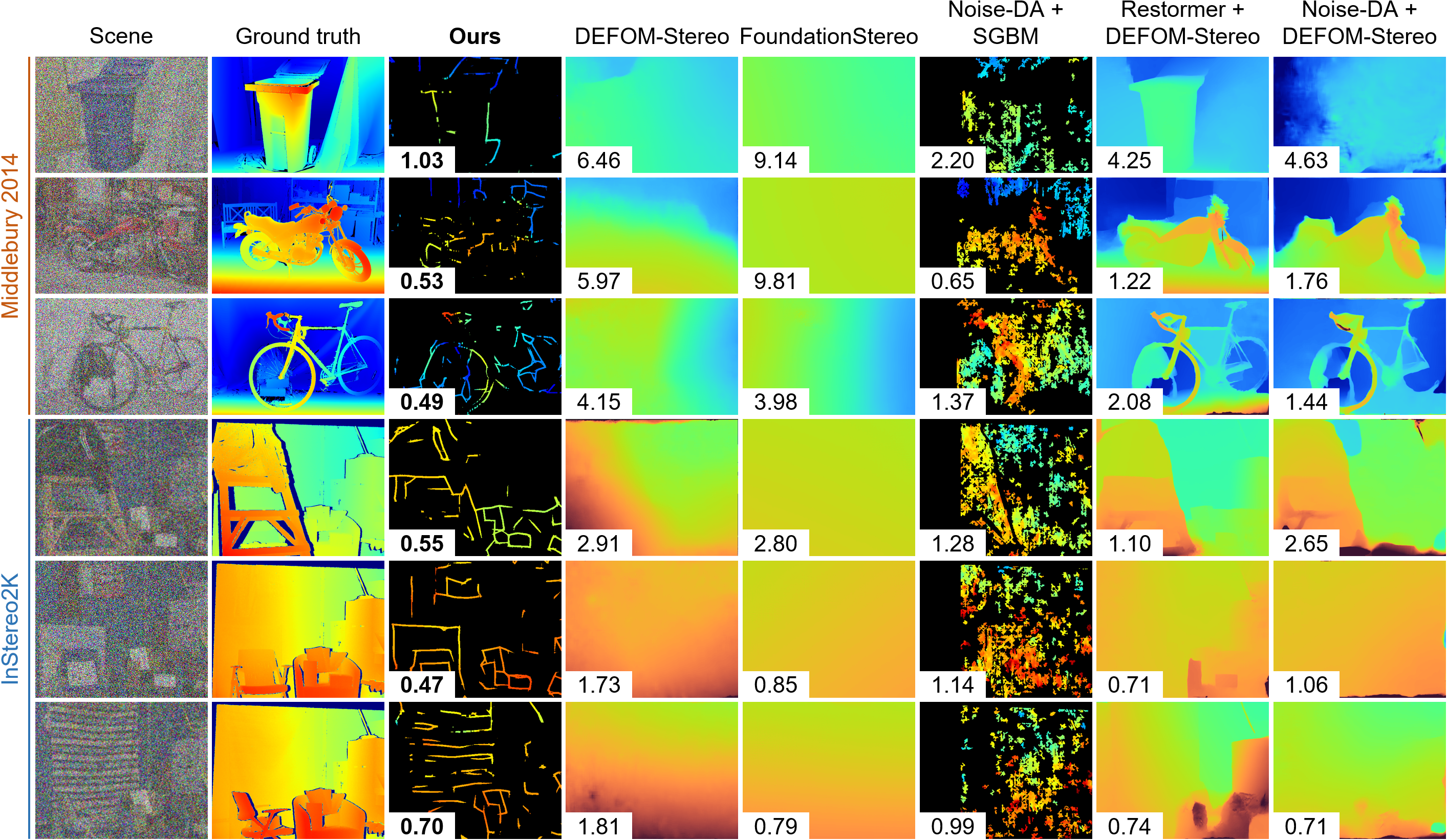}
    \caption{Qualitative comparison of disparity estimation on the Middlebury 2014~\cite{scharstein2014high} and InStereo2K~\cite{bao2020instereo2k} datasets. Disparity maps are visualized using per-image color normalization derived from the ground truth (GT) disparity range; predicted and GT disparities are clamped to $[0.9\times \min_{\text{GT}}, 1.1\times \max_{\text{GT}}]$. Each row shows the left input image, GT disparity, and predictions from competing methods, with EPE~\cite{scharstein2002taxonomy} (px) reported for each method.}
    \label{fig:qualitative_comparison}
\end{figure*}

\paragraph*{Qualitative Analysis.}
Figure~\ref{fig:qualitative_comparison} shows that our method reliably recovers discrete edge structures and produces accurate disparity estimations along these structures from severely corrupted input image pairs. Denoising-based pipelines also demonstrate competitive performance: Restormer combined with DEFOM-Stereo and Noise-DA combined with SGBM yield lower accuracy but dense disparity estimations. In the supplement, we show that our sparse disparity map can be readily densified using off-the-shelf, foundation densification models, e.g.,~\cite{viola2024marigolddc}, while still achieving the best performance on most scenes from Middlebury 2014. 

\section{Discussion}

The proposed non-learning framework demonstrates a promising direction for low-light stereo vision. Rather than attempting to restore corrupted image features, our approach first extracts coarse visual features that remain robust under noisy conditions. Disparity matching is then performed using only these features, yielding a sparse disparity map. These sparse predictions are shown to be highly accurate, with an optional downstream procedure available for densification.

Current limitations include the high computational cost of extracting the FoJ representations and the lack of joint regularization between the structures of the two stereo images. Future work may explore a unified framework that integrates structure recovery and disparity estimation in a more coherent manner.

\section{Acknowledgement}

This work was partially supported by the National Science Foundation (NSF) under Grant No. CCF-2431505 and CCF-2544069.

\bibliographystyle{IEEEbib}
{
\small
\bibliography{refs}
}

\clearpage

\twocolumn[
\begin{center}
\textbf{\fontsize{12}{14.4}\selectfont NON-LEARNING LOW-LIGHT STEREO VISION}\\
\vspace{0.1in}
\fontsize{12}{14.4}\selectfont SUPPLEMENTARY MATERIAL
\vspace{0.3in}
\end{center}
]

\setcounter{section}{0}
\setcounter{figure}{0}
\setcounter{table}{0}
\setcounter{equation}{0}

\renewcommand{\thesection}{S\arabic{section}}
\renewcommand{\thefigure}{S\arabic{figure}}
\renewcommand{\thetable}{S\arabic{table}}
\renewcommand{\theequation}{S\arabic{equation}}

\section{Window Size Sweep}

\begin{figure*}[tb]
    \centering
    \includegraphics[width=\linewidth]{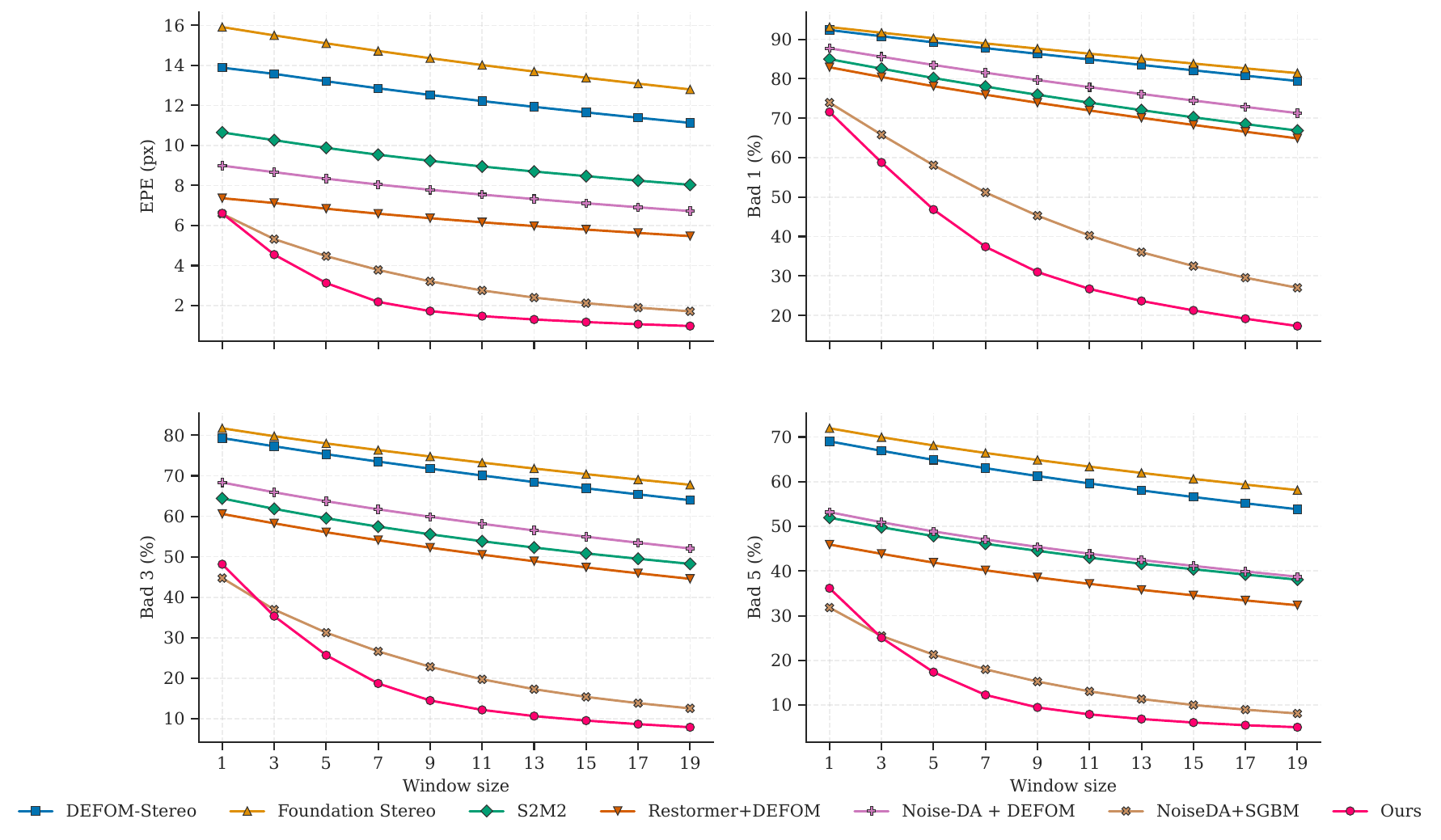}
    \captionof{figure}{EPE (px) and Bad-\textit{K} rates for $K = 1, 3, 5$ (\%) as the evaluation window size $W$ is varied from $1$ to $19$ on Middlebury 2014 ($\alpha=2$)~\cite{scharstein2014high}. $W =1$ is the standard pointwise metric, larger $W$ relaxes spatial tolerance. We compare our method against the top-performing baselines.}
    \label{fig:window_sweep}
\end{figure*}

In the main paper, we report metrics evaluated with a fixed window size of $W=19$ px. Here we examine how metrics (EPE and Bad-\textit{K}) change as $W$ is swept from 1 to 19 px. Figure~\ref{fig:window_sweep} shows that our method's error drops sharply with $W$ and quickly plateaus, while curves for SOTA methods appear roughly linear. Noise-DA~\cite{liao2024denoising} + SGBM~\cite{hirschmuller2007stereo} shows a decay shape similar to that of our method; however, it appears flatter and plateaus at a higher value than ours.

\paragraph*{Interpretation.} Curves that drop sharply indicate methods whose metrics are dominated by small localization errors rather than inaccurate disparity estimates. Linear curves indicate methods whose errors persist under spatial tolerance, i.e., methods producing structurally incorrect predictions. Our method consistently maintains its advantage across all $W \geq 3$ and shows a clear, sharp decay and plateau, consistent with errors dominated by localization offsets. This consistent advantage and decay shape also indicate that the improvements reported in the main paper are not an artifact of evaluation tolerance.

These results are consistent with the hypothesis that our method's predictions are accurate in value but spatially displaced from true boundaries, whereas SOTA methods have precise localization but poor absolute accuracy, and thus gain little additional benefit from a larger window.

\section{Depth Completion}
\paragraph*{Pipeline Details.}
As our method produces sparse disparity estimates, we densify the output using a depth completion model, Marigold-DC~\cite{viola2024marigolddc}. The left image is first denoised with Restormer~\cite{Zamir2021Restormer} to provide a clean reference image for Marigold-DC. We then run our sparse stereo method on noisy input to obtain a sparse disparity map. The disparity map is then converted to a sparse depth map via $z = 1 / \hat{d}$. Both the sparse depth map and the denoised left image are fed into Marigold-DC, and the resulting dense depth is inverted to produce the dense disparity map.

We use the official Marigold-DC implementation\footnote{\url{https://github.com/prs-eth/Marigold-DC}}
with the following arguments:
\begin{itemize}
    \itemsep0em
    \item \texttt{num\_inference\_steps}: 50
    \item \texttt{processing\_resolution}: 0 (native resolution)
    \item \texttt{use\_full\_precision}: enabled (fp32)
    \item \texttt{ensemble\_size}: 10
\end{itemize}
All other parameters are left at their default values.

\begin{figure*}[h!]
    \centering
    \includegraphics[width=.875\textwidth]{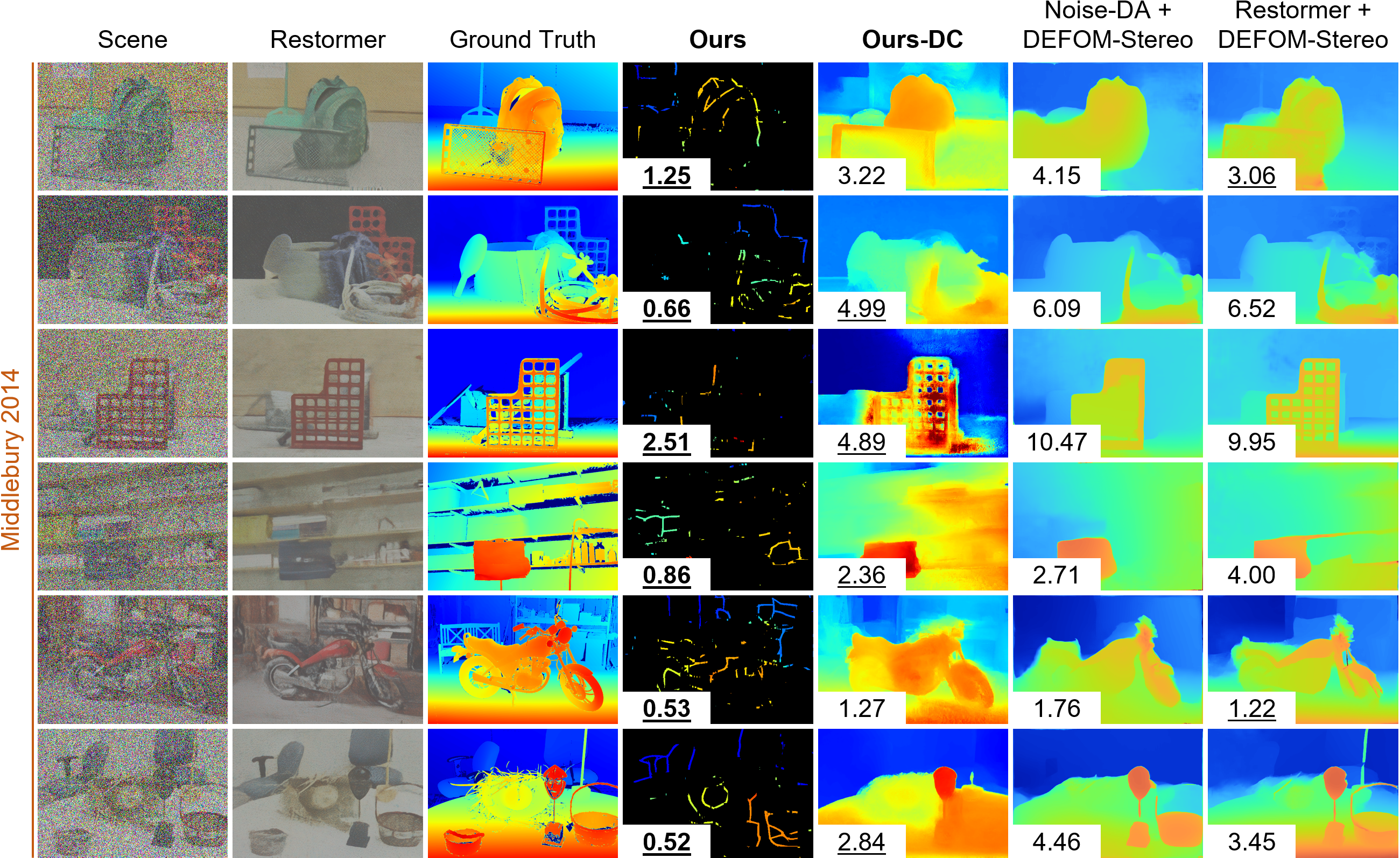}
    \caption{Qualitative comparison of depth completion results on Middlebury 2014~\cite{scharstein2014high}. Disparity maps are visualized using per-image color normalization derived from the ground truth (GT) disparity range; predicted and GT disparities are clamped to $[0.9\times \min_{\text{GT}}, 1.1\times \max_{\text{GT}}]$. Each row shows the left input image, Restormer-denoised~\cite{Zamir2021Restormer} image, GT disparity, our sparse prediction, and predictions from the three dense methods, with EPE~\cite{scharstein2002taxonomy} (px) reported for each method.}
    \label{fig:dc_qual_comp}
\end{figure*}

\paragraph*{Qualitative Comparison.}
We compare our framework, extended with depth completion (Ours-DC), with two top-performing dense SOTA methods:  Restormer~\cite{Zamir2021Restormer} + DEFOM-Stereo~\cite{jiang2025defom}, and Noise-DA~\cite{liao2024denoising} + DEFOM-Stereo~\cite{jiang2025defom}. Fig.~\ref{fig:dc_qual_comp} shows that Ours-DC is competitive with dense SOTA in EPE on the scenes shown, never falling more than $0.16$ px behind, while producing visual quality that approaches SOTA quality. We view densification of our sparse predictions as a promising direction for future work.

\section{Ablation Study}

To validate the efficacy of our novel contributions, we ablate our boundary-aware SGM formulation and symmetric cost function. All methods reported in Tab.~\ref{tab:ablation} are evaluated on Middlebury 2014 ($\alpha=2$)~\cite{scharstein2014high}. 

\paragraph*{Boundary-Aware SGM Ablation.}
To assess the contribution of our boundary-aware SGM formulation, we compare our full method against a variant that uses the standard SGM~\cite{hirschmuller2007stereo} algorithm with fixed $P_2=P_{2,\text{base}}=200$, matching the base value used in our full method. Table~\ref{tab:ablation} shows that the ablated variant trails behind the full method by around $0.4$ px in EPE and 3--4\% on the Bad-\textit{K} metrics. The higher Valid (\%) of the fixed-$P_2$ variant reflects over-smoothing across boundaries that our boundary-aware modulation suppresses.

\paragraph*{Symmetric Cost Function Ablation.}
To isolate the contribution of the bidirectional matching, we compare against a variant that replaces $\mathbf{C}_L, \mathbf{C}_R$ with the standard non-symmetric form, retaining only the reconstruction-to-raw term in each direction:
\begin{equation}
    \mathbf{C}_{L/R}(\x_i, d) = \lVert C_{L/R}(\x_i) - P_{R/L}(\x_i \pm d) \rVert^2.
\end{equation}

Table~\ref{tab:ablation} shows the ablated variant trails behind the full method by around $1.6$ px in EPE, and 7--9\% on the Bad-\textit{K} metrics. The lower Valid (\%) reflects that fewer pixels pass left-right consistency without the symmetric cross-terms.

\begin{table*}[h!]
\centering
\caption{Ablation on Middlebury 2014 ($\alpha=2$), comparing our full method against two ablated variants: a fixed SGM $P_2$ penalty without boundary weighting, and a classical non-symmetric cost. We report EPE and Bad-\textit{K} metrics evaluated with $W=19$. Valid (\%) denotes the fraction of unmasked pixels.}
\label{tab:ablation}
\begin{tabular}{@{}l|cccc|c@{}}
\hline
\textbf{Method} & \textbf{EPE (px)} $\downarrow$ & \textbf{Bad-1 (\%)} $\downarrow$ & \textbf{Bad-3 (\%)} $\downarrow$ & \textbf{Bad-5 (\%)} $\downarrow$ & \textbf{Valid (\%)} \\
\hline

\textbf{Ours} & \textbf{0.9782} & \textbf{17.30} & \textbf{7.88}  & \textbf{5.03}  & 2.82 \\
Fixed SGM Penalty (no boundary weights) & 1.3808 & 21.13 & 11.65 & 8.09 & 3.70 \\
Classical (non-symmetric) cost & 2.6058 & 27.06 & 15.99 & 12.21 & 0.56 \\
\hline
\end{tabular}
\end{table*}
\end{document}